\documentclass[10pt,twocolumn,letterpaper]{article}

\usepackage[article]{cvpr}
\usepackage{graphicx}
\usepackage{float}
\usepackage{amsmath}
\usepackage{amssymb}
\usepackage{booktabs}
\usepackage{ctable}
\usepackage{multirow}
\usepackage{array}
\usepackage{colortbl}

\definecolor{cvprblue}{rgb}{0.21,0.49,0.74}
\usepackage[pagebackref,breaklinks,colorlinks,citecolor=cvprblue]{hyperref}

\def\paperID{}
\def\confName{CVPR}
\def\confYear{}

\begin{document}

\title{Systematic Evaluation of Learning Rate Scheduling Strategies Across Heterogeneous Architectures}

\author{Hafsa Mateen{\thanks{Corresponding author: hafsa.mateen@stud-mail.uni-wuerzburg.de}},
        \space\space\space Radu Timofte,
        \space\space\space Dmitry Ignatov\\
    \small{Computer Vision Lab, CAIDAS \& IFI, University of W\"urzburg, Germany}}
\maketitle
\pagestyle{plain}

\begin{abstract}
Choosing a learning rate scheduling strategy is critical to neural network training, but manual selection is costly and rarely exhaustive. While classical AutoML approaches often treat the scheduler as a secondary hyperparameter, we systematically investigate its impact on classification accuracy across a diverse pool of architectures.

We evaluated 30 representative architectures from convolutional and transformer families within the LEMUR neural network dataset. Through automated source-code injection, we applied 25 scheduler configurations from ten PyTorch scheduler families, evaluating a total of 3,938 model variants on CIFAR-10 for five epochs each. Top-1 accuracy ranges from 12.08\% to 86.45\% with a mean of 52.52\%, and 237 variants exceed 80\%. Mean accuracy per scheduler family spans 45.04\% to 55.20\%, whereas the best result per family varies by only about two percentage points. The best scheduler differs between architectures: the best runs of the 20 individually attributed architectures are spread over all ten families. The resulting accuracy landscape, contributed to the LEMUR nn-dataset, is a reference for scheduler selection in short-budget screening.
\end{abstract}

\section{Introduction}
\label{sec:intro}

Deep learning has witnessed remarkable progress over the past decade,
driven by advances in model architectures, large-scale datasets, and
computational infrastructure. Yet the training recipe --- the precise
combination of optimizer, learning rate schedule, regularization, and
augmentation strategy --- remains a largely hand-crafted artifact that
requires substantial expert knowledge and empirical trial.

Among the components of a training recipe, the learning rate (LR)
schedule exerts a particularly strong influence on convergence speed,
final accuracy, and generalization. A poorly chosen schedule can cause
slow convergence, oscillation around the optimum, or premature
saturation, while a well-chosen one can recover several percentage
points of accuracy with no change to the model architecture. Despite
this significance, most published works report results with a single
fixed schedule and treat the scheduler as a secondary hyperparameter.

Classical hyperparameter optimization (HPO) methods such as Bayesian
optimization~\cite{snoek2012bayesian} and random search can in principle
explore the scheduling space, but they are expensive, require repeated
full training runs, and offer limited interpretability. Systematic
grid-based evaluation across a diverse architecture pool offers a
complementary and reproducible alternative: by fixing all other
training hyperparameters and varying only the scheduler, one can
directly attribute accuracy differences to scheduling choices and
build a reusable reference landscape.

We present such a study, contributing a large-scale systematic
evaluation of learning rate scheduling strategies across 30
heterogeneous neural network architectures evaluated on CIFAR-10.

Our contributions are as follows.

\begin{itemize}
    \item We contribute 3,938 LR-scheduler model variants to the LEMUR
    nn-dataset~\cite{ABrain.NN-Dataset, ABrain.LEMUR2, ABrain.NN-Lite}
    by injecting 25 diverse LR scheduler configurations into 30 neural
    network architectures via automated source-code injection,
    evaluating all 3,938 variants on CIFAR-10 for five
    epochs.

    \item We systematically evaluate the impact of scheduler choice on
    top-1 accuracy. Scheduler family shifts mean five-epoch accuracy by
    about ten percentage points but the best reachable accuracy by only
    about two, and the best scheduler differs from architecture to
    architecture.

    \item We report accuracy as a function of weight decay and the best
    accuracy and scheduler for each individually attributed
    architecture.

    \item The best configuration achieves 86.45\% top-1 accuracy, and
    237 of 3,938 variants exceed 80\%.
\end{itemize}

The remainder of this paper is organized as follows.
Section~\ref{sec:Related} reviews related work.
Section~\ref{sec:Methodology} describes the architecture pool,
scheduler catalogue, and injection pipeline.
Section~\ref{sec:experiments} details the experimental setup.
Section~\ref{sec:results} presents results and analysis.
Section~\ref{sec:limitations} discusses limitations.
Section~\ref{sec:conclusion} concludes.

\section{Related Work}
\label{sec:Related}

\subsection{Learning Rate Scheduling}

Learning rate scheduling has a long history in deep learning
optimization. Fixed-step decay was among
the earliest systematic approaches, reducing the learning rate by a
multiplicative factor at predetermined epochs. Cosine
annealing~\cite{loshchilov2017sgdr} introduced a smooth decay
along a cosine curve, optionally with periodic warm restarts, and is
widely used for training vision models. The one-cycle
policy~\cite{smith2019super} takes a fundamentally different
approach, first warming up the learning rate to a maximum before
annealing, achieving super-convergence on several benchmarks.
Cyclical learning rates~\cite{smith2017cyclical} vary the rate
cyclically between a lower and an upper bound instead of decreasing it
monotonically. Adaptive methods such as
ReduceLROnPlateau monitor a validation metric and reduce the rate
when progress stalls, providing a data-driven alternative to fixed
schedules.

Warm-up strategies, in which the learning rate is gradually increased
from a small value at the start of training, are commonly used when
training large vision transformers, often followed by cosine annealing.
Despite the proliferation of
scheduling methods, systematic large-scale comparisons across diverse
architecture families remain rare, and most published results report
a single fixed schedule without ablating the scheduler choice.

\subsection{Automated Hyperparameter Optimization}

Classical AutoML approaches such as Bayesian optimization,
evolutionary search, and random search have been widely applied to
hyperparameter tuning. Kochnev \emph{et al.}~\cite{ABrain.HPGPT}
fine-tuned Code Llama to recommend hyperparameters for vision models
and reported results competitive with or better than Optuna's
Tree-structured Parzen Estimator at lower computational cost.
Vysyaraju \emph{et al.}~\cite{ABrain.Prompt} studied few-shot
prompting and efficient validation for LLM-based neural network
generation. Gu \emph{et
al.}~\cite{ABrain.Feedback_Memory} added a feedback memory of recent
improvement attempts to an iterative LLM-based neural architecture
search loop. These works collectively demonstrate that
LLMs can reduce the cost of hyperparameter search while maintaining
competitive performance, motivating systematic evaluation of the
scheduling dimension across diverse architecture families.

\subsection{Automated Neural Network Pipelines}

Recent work within the NNGPT framework~\cite{ABrain.NNGPT} has
demonstrated automated generation and evaluation of neural network
implementations across diverse architecture families. Supporting
studies have explored architecture design~\cite{ABrain.Architect},
data transformation~\cite{ABrain.Transform}, and retrieval-based
analysis of neural network behaviour~\cite{ABrain.NN-RAG} within
such pipelines. A reliable training recipe --- including a
well-chosen learning rate scheduler --- is a prerequisite for
meaningful evaluation scores in any automated pipeline. Our work
directly addresses this gap by providing a systematic empirical
reference for scheduler selection across the heterogeneous
architectures used in these frameworks.

\subsection{Neural Network Datasets}

The LEMUR neural network dataset~\cite{ABrain.NN-Dataset} is an
open-source collection of PyTorch neural networks that follow a unified
template, which allows architectures to be compared under the same
training code. LEMUR 2~\cite{ABrain.LEMUR2} extends the dataset to more
than 14,000 architectures and adds further tasks and on-device latency
benchmarks. Din \emph{et al.}~\cite{ABrain.NN-Lite} describe an
automated pipeline for deploying and benchmarking PyTorch models on
Android devices. These datasets form the foundation of our experimental
pool and allow scheduler performance to be evaluated across a broad set
of architectures within one evaluation framework.

\subsection{CIFAR-10 as a Benchmark}

CIFAR-10~\cite{krizhevsky2009} is a widely used image
classification benchmark consisting of 60,000 $32\!\times\!32$ colour
images across 10 classes, split into 50,000 training and 10,000 test
samples. Its compact size makes it well-suited for large-scale
hyperparameter sweeps where full-resolution datasets such as
ImageNet would be prohibitively expensive. Prior work on learning
rate scheduling~\cite{loshchilov2017sgdr, smith2017cyclical, smith2019super}
also reports results on CIFAR-10. Because our runs use a five-epoch
budget and inputs resized to $256\!\times\!256$, absolute accuracies are
not comparable to fully trained results reported in those works, and
whether the rankings transfer to other datasets is not tested here.

\section{Methodology}
\label{sec:Methodology}

Our methodology consists of two stages: (i) systematic construction
of a large-scale model variant pool by injecting diverse LR scheduling
strategies into existing neural network architectures, and (ii) automated
evaluation of all generated variants using a unified training
pipeline.

\subsection{Architecture Pool and the LEMUR Dataset}
\label{sec:arch_pool}

To ensure broad coverage across architecture families, we build upon
the LEMUR dataset~\cite{ABrain.NN-Dataset, ABrain.LEMUR2,
ABrain.NN-Lite}, which provides a diverse collection of neural
network implementations sharing a unified training interface. From
this collection we select 30 representative architectures spanning
convolutional and transformer-based families:
AlexNet, ResNet, VGG, DenseNet, EfficientNet, MobileNetV2/V3,
RegNet, ShuffleNet, SqueezeNet, GoogLeNet, InceptionV3, MNASNet,
MaxVit, SwinTransformer, VisionTransformer, ConvNeXt,
DPN68/107/131, BagNet, FractalNet, AirNet, AirNext, BayesianNet,
DarkNet, UNet2D, ICNet, Diffuser, and ComplexNet. Each architecture
exposes a standardised \texttt{train\_setup()} method for optimiser
and scheduler initialisation and a \texttt{learn()} method for the
per-iteration training loop, enabling fully automated code injection.
The selection covers lightweight mobile-optimized models, standard
mid-capacity convolutional networks, dense connection networks,
and vision transformers.

\subsection{Learning Rate Scheduler Catalogue}
\label{sec:scheduler_catalogue}

We define a catalogue of 25 LR scheduler configurations drawn from
ten PyTorch~\cite{paszke2019pytorch} scheduler families, summarised in
Table~\ref{table:schedulers}.

\textbf{StepLR} (4 variants): decays by $\gamma$ every fixed step
fraction $\{0.10, 0.25, 0.50, 0.70\}$ of the budget, with
$\gamma \in \{0.1, 0.3, 0.5, 0.7\}$.

\textbf{ExponentialLR} (3 variants): per-epoch decay
$\text{lr}_{t} = \text{lr}_0 \cdot \gamma^{t}$,
$\gamma \in \{0.90, 0.95, 0.98\}$.

\textbf{CosineAnnealingLR} (3 variants): cosine decay with
$\eta_{\min} = 10^{-6}$ and $T_{\max} \in \{5, 10, 20\}$ epochs.

\textbf{CosineAnnealingWarmRestarts} (2 variants): cosine with
periodic restarts, $T_0 \in \{2, 5\}$ epochs.

\textbf{MultiStepLR} (3 variants): milestones
$\{5,10\}$, $\{3,7\}$, $\{2,4\}$ with $\gamma \in \{0.5,0.3,0.1\}$.

\textbf{ReduceLROnPlateau} (3 variants): factor--patience
$(0.5,2)$, $(0.3,3)$, $(0.1,5)$.

\textbf{CyclicLR} (3 variants): \emph{triangular},
\emph{triangular2}, \emph{exp\_range} between
$10^{-4}$ and $0.1$.

\textbf{OneCycleLR} (2 variants):
$\text{lr}_{\max} \in \{0.05, 0.10\}$.

\textbf{LinearLR} and \textbf{PolynomialLR} (1 each):
fixed defaults (start factor 0.1 over 5 epochs; degree-2 decay
over 5 iterations, respectively).

\begin{table}[t]
    \caption{Summary of the 25 LR scheduler configurations.}
    \label{table:schedulers}
    \centering
    \fontsize{6.5}{8}\selectfont
    \begin{tabular}{lp{3.0cm}c}
        \toprule
        Family & Key Parameters & \# \\
        \midrule
        StepLR & step$\in$\{0.1,0.25,0.5,0.7\}, $\gamma\in$\{0.1,0.3,0.5,0.7\} & 4 \\
        ExponentialLR & $\gamma\in$\{0.90,0.95,0.98\} & 3 \\
        CosineAnnealingLR & $T_{\max}\in$\{5,10,20\} epochs & 3 \\
        CosineAnnealingWR & $T_0\in$\{2,5\} & 2 \\
        MultiStepLR & milestones \{5,10\},\{3,7\},\{2,4\} & 3 \\
        ReduceLROnPlateau & (factor,patience): (0.5,2),(0.3,3),(0.1,5) & 3 \\
        CyclicLR & triangular, triangular2, exp\_range & 3 \\
        OneCycleLR & $\text{lr}_{\max}\in$\{0.05,0.10\} & 2 \\
        LinearLR & fixed defaults & 1 \\
        PolynomialLR & fixed defaults & 1 \\
        \midrule
        \textbf{Total} & & \textbf{25} \\
        \bottomrule
    \end{tabular}
\end{table}

\subsection{Automated Code Injection}
\label{sec:injection}

For each architecture--scheduler--weight\_decay combination we
programmatically generate a modified model variant using the
\texttt{inject\_scheduler()} function. This function performs
line-level source-code editing guided by Python parsing in four
steps.

\textbf{Step 1 --- Method boundary detection.} Locates
\texttt{train\_setup()} and \texttt{learn()} within \texttt{class Net}
by tracking indentation levels, ensuring robustness across diverse
coding styles present in the LEMUR dataset.

\textbf{Step 2 --- Scheduler initialisation injection.} Scheduler
instantiation code is appended to \texttt{train\_setup()}. Where a
scheduler already exists, it is replaced rather than duplicated.

\textbf{Step 3 --- Scheduler step injection.} For per-epoch schedulers
(\emph{e.g.}, StepLR, CosineAnnealingLR),
\texttt{self.scheduler.step()} is inserted at the end of
\texttt{learn()}. For per-batch schedulers (\emph{e.g.}, CyclicLR,
OneCycleLR, ReduceLROnPlateau), the step call is placed immediately
after \texttt{self.optimizer.step()}. ReduceLROnPlateau is
therefore stepped once per batch on the mini-batch training loss, so
its patience is counted in batches rather than epochs.

\textbf{Step 4 --- Hyperparameter registration.} The
\texttt{supported\_hyperparameters()} function is rewritten to
register all scheduler-specific parameters alongside the original
architecture hyperparameters, ensuring full compatibility with the
NNEval evaluation harness.

Each variant is validated by (i) parsing the modified source with
Python's \texttt{ast} module to verify syntactic correctness, and
(ii) confirming every registered hyperparameter key appears at least
twice as a string literal in the code. Variants failing either check
are discarded. A bug in \texttt{LinearLR} and \texttt{PolynomialLR}
--- where \texttt{total\_iters} was incorrectly scaled, causing step
count mismatch --- was identified and fixed prior to the main
evaluation run; the 516 results of the second evaluation phase were
produced after this fix (Section~\ref{sec:train_test}).

\subsection{Hyperparameter Search Space}
\label{sec:hp_space}

Beyond scheduler type, we vary weight decay across seven values:
$\lambda \in \{0, 10^{-5}, 5\!\times\!10^{-5}, 10^{-4},
5\!\times\!10^{-4}, 10^{-3}, 5\!\times\!10^{-3}\}$.
All other hyperparameters are fixed:
$\text{lr} = 0.01$, batch size $B = 64$, dropout $p = 0.2$,
momentum $\mu = 0.9$, $E_{\max} = 5$ epochs,
\texttt{norm\_256\_flip} augmentation. The full combinatorial space
yields $30 \times 25 \times 7 = 5{,}250$ theoretical variants;
those passing syntax validation and hyperparameter consistency checks
are retained for evaluation. Weight decay was included as an
additional dimension because it interacts strongly with learning rate
magnitude and schedule shape, and its effect is expected to vary
across architecture families.

\section{Experiments}
\label{sec:experiments}

\subsection{Dataset}
\label{sec:dataset}

All experiments use the \textbf{CIFAR-10} dataset, which consists of
60,000 colour images of size $32\!\times\!32$ pixels across 10 object
classes (airplane, automobile, bird, cat, deer, dog, frog, horse,
ship, truck), split into 50,000 training and 10,000 test images.
The \texttt{norm\_256\_flip} transform is applied uniformly: images
are resized to $256\!\times\!256$, normalised using per-channel mean
and standard deviation, and randomly horizontally flipped during
training. CIFAR-10 was chosen because its compact size allows
evaluation of nearly 4,000 model variants within a reasonable
computational budget while remaining a widely used and well-understood
benchmark for image classification.

\subsection{Training and Testing}
\label{sec:train_test}

All model variants are trained and evaluated using the NNEval harness
on NVIDIA GeForce RTX 3090/4090 (24\,GB) GPUs within the CVL
Kubernetes cluster at the University of W\"urzburg. Each model is
trained for five epochs in the screening phase to efficiently identify
promising scheduler--architecture combinations before committing to
full-length evaluation.

All models use SGD with $\text{lr} = 0.01$, momentum $\mu = 0.9$,
batch size $B = 64$, and dropout $p = 0.2$. Architecture-specific
parameters such as stochastic depth probability for SwinTransformer
and ConvNeXt, and attention dropout for VisionTransformer, are set
to their recommended defaults throughout. The evaluation was
conducted across multiple phases: an initial run (March 26, 2026)
producing 1,507 model results, a second run (April 24 --
May 2, 2026) adding 516 models after a bug fix in the LinearLR
and PolynomialLR scheduler injection code, and a third phase
extending coverage to additional architecture--scheduler combinations,
yielding 3,938 total evaluated variants contributed to the LEMUR
nn-dataset.

The five-epoch screening budget was chosen to keep the evaluation
of nearly 4,000 variants affordable. It rests on the assumption that
accuracy after a short budget is informative about longer training, an
assumption that early-stopping methods such as
Hyperband~\cite{li2018hyperband} also make; we do not verify it here
(Section~\ref{sec:limitations}). The hardware achieved
approximately 200 samples per second per worker, enabling the full
3,938-variant evaluation to complete within a practical timeframe.

\subsection{Evaluation Metrics}
\label{sec:metrics}

The primary metric is \textbf{top-1 classification accuracy} on the
CIFAR-10 test set. We additionally report: (i) the number of variants
exceeding 75\% and 80\% accuracy thresholds to characterise the upper
tail of the performance distribution; (ii) mean and median accuracy
to measure overall landscape quality; (iii) per-scheduler-family mean
and best accuracy to identify the most effective scheduling families;
(iv) mean and best accuracy per weight-decay value; and (v)
per-architecture best accuracy and the scheduler family of that run.

\section{Results and Discussion}
\label{sec:results}

\subsection{Overall Performance}

Across 3,938 evaluated model variants, top-1 accuracy ranges from
12.08\% to 86.45\%, with a mean of 52.52\% and a median of 50.00\%.
The spread of 74.37 percentage points covers the whole grid of
architectures, scheduler configurations and weight-decay values, so it
cannot be attributed to the scheduler alone; the scheduler's share is
examined in Section~\ref{sec:per_scheduler}. A total of 237 variants
(6.0\%) exceed 80\% accuracy, 627 variants (15.9\%) exceed 75\%, and
1,107 variants (28.1\%) exceed 60\% (Table~\ref{table:stats}).

\begin{table}[t]
    \caption{Summary statistics across all 3,938 evaluated variants.}
    \label{table:stats}
    \centering
    \fontsize{7.5}{9}\selectfont
    \begin{tabular}{lc}
        \toprule
        Metric & Value \\
        \midrule
        Total variants contributed \& evaluated & 3,938 \\
        Maximum accuracy            & 86.45\% \\
        Mean accuracy               & 52.52\% \\
        Median accuracy             & 50.00\% \\
        Minimum accuracy            & 12.08\% \\
        Variants $>$ 80\% accuracy  & 237 (6.0\%) \\
        Variants $>$ 75\% accuracy  & 627 (15.9\%) \\
        Variants $>$ 60\% accuracy  & 1,107 (28.1\%) \\
        Variants $<$ 30\% accuracy  & 445 (11.3\%) \\
        \bottomrule
    \end{tabular}
\end{table}

\subsection{Per-Scheduler Analysis}
\label{sec:per_scheduler}

\begin{figure}[t]
    \centering
    \includegraphics[width=\linewidth]{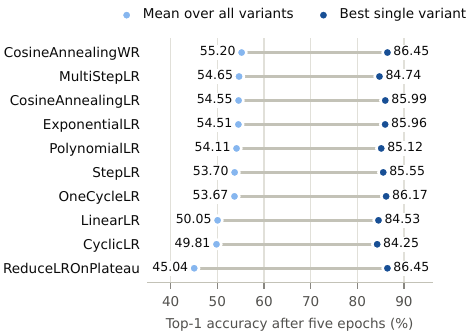}
    \caption{Mean and best top-1 accuracy per scheduler family over all
    3,938 variants (values from Table~\ref{table:per_scheduler}). Family
    means differ by about ten percentage points, the best results by
    about two.}
    \label{fig:scheduler_mean}
\end{figure}

Table~\ref{table:per_scheduler} and Figure~\ref{fig:scheduler_mean}
report mean and best accuracy per scheduler family. Mean accuracy per
family spans 45.04\% to 55.20\%, a range of about ten percentage
points, whereas the best single result per family lies between 84.25\%
and 86.45\%, a range of about two points. Scheduler family therefore
mainly shifts the average outcome of a five-epoch run, not the best
accuracy that can be reached.

CosineAnnealingWarmRestarts has the highest mean (55.20\%) and shares
the best single result (86.45\%) with ReduceLROnPlateau. Seven families
lie within about 1.5 points of each other (53.67--55.20\% mean), so
their order should not be over-interpreted. LinearLR (50.05\%),
CyclicLR (49.81\%) and ReduceLROnPlateau (45.04\%) lag behind.
ReduceLROnPlateau has the lowest mean but still reaches 86.45\% in its
best configuration. Because it is stepped per batch on the mini-batch
loss (Section~\ref{sec:injection}), its patience of 2--5 is counted in
batches, so the rate can be reduced early and repeatedly; this is the
likely reason for its low mean.

\begin{table}[t]
    \caption{Mean and best top-1 accuracy per scheduler family.}
    \label{table:per_scheduler}
    \centering
    \fontsize{7.5}{9}\selectfont
    \begin{tabular}{lcc}
        \toprule
        Scheduler Family & Mean Acc. (\%) & Best Acc. (\%) \\
        \midrule
        \textbf{CosineAnnealingWR}   & \textbf{55.20} & \textbf{86.45} \\
        MultiStepLR         & 54.65 & 84.74 \\
        CosineAnnealingLR   & 54.55 & 85.99 \\
        ExponentialLR       & 54.51 & 85.96 \\
        PolynomialLR        & 54.11 & 85.12 \\
        StepLR              & 53.70 & 85.55 \\
        OneCycleLR          & 53.67 & 86.17 \\
        LinearLR            & 50.05 & 84.53 \\
        CyclicLR            & 49.81 & 84.25 \\
        ReduceLROnPlateau   & 45.04 & \textbf{86.45} \\
        \bottomrule
    \end{tabular}
\end{table}

\subsection{Weight Decay Analysis}

\begin{figure}[t]
    \centering
    \includegraphics[width=\linewidth]{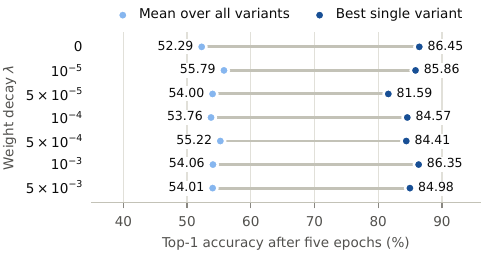}
    \caption{Mean and best top-1 accuracy per weight-decay value
    $\lambda$, aggregated across all architectures and schedulers
    (values from Table~\ref{table:weight_decay}).}
    \label{fig:weight_decay}
\end{figure}

Table~\ref{table:weight_decay} and Figure~\ref{fig:weight_decay} report mean and best accuracy as a
function of weight decay value, aggregated across all architectures
and scheduler types. $\lambda = 10^{-5}$ yields the highest mean
(55.79\%), while $\lambda = 0$ achieves the highest peak (86.45\%).
The means of all six non-zero values (53.76--55.79\%) lie above the
mean at $\lambda = 0$ (52.29\%); the largest value,
$\lambda = 5\!\times\!10^{-3}$, has a mean of 54.01\%. The best
accuracy at each weight-decay value lies between 81.59\% and 86.45\%.
Because the numbers of variants per weight-decay value are not equal
and no significance test was performed, these differences are
observations rather than evidence of an effect of weight decay.

\begin{table}[t]
    \caption{Mean and best top-1 accuracy as a function of weight
    decay $\lambda$, aggregated across all architectures and
    schedulers.}
    \label{table:weight_decay}
    \centering
    \fontsize{7.5}{9}\selectfont
    \begin{tabular}{ccc}
        \toprule
        Weight Decay $\lambda$ & Mean Acc. (\%) & Best Acc. (\%) \\
        \midrule
        $0$                    & 52.29 & \textbf{86.45} \\
        $\mathbf{10^{-5}}$              & \textbf{55.79} & 85.86 \\
        $5\!\times\!10^{-5}$  & 54.00 & 81.59 \\
        $10^{-4}$              & 53.76 & 84.57 \\
        $5\!\times\!10^{-4}$  & 55.22 & 84.41 \\
        $10^{-3}$              & 54.06 & 86.35 \\
        $5\!\times\!10^{-3}$  & 54.01 & 84.98 \\
        \bottomrule
    \end{tabular}
\end{table}

\subsection{Per-Architecture Analysis}

Table~\ref{table:per_arch} reports best accuracy and best scheduler
per architecture, sorted by best accuracy. Ten of the 30 pool entries could not be
individually attributed due to class-naming conventions in the source
files: AirNext, ComplexNet, MNASNet, MaxVit, VGG, AlexNet, ShuffleNet,
and Diffuser, together with two of the three DPN variants
(DPN68/107/131), which are reported under a single DPN row. They are
excluded from the per-architecture table; their results are included in
the aggregate statistics.
GoogLeNet and InceptionV3 jointly achieve the highest peak (86.45\%),
with GoogLeNet under CosineAnnealingWarmRestarts and InceptionV3
under ReduceLROnPlateau. DenseNet (86.17\%) and RegNet (85.96\%)
follow closely. The 20 best results are spread over all ten scheduler
families, and no family is best for more than three architectures. Each
entry is the maximum over all runs of that architecture, so it shows
where the best run occurred rather than an average preference.

\begin{table}[t]
    \caption{Best top-1 accuracy and best scheduler per architecture.}
    \label{table:per_arch}
    \centering
    \fontsize{6.5}{8}\selectfont
    \begin{tabular}{lcc}
        \toprule
        Architecture & Best Acc. (\%) & Best Scheduler \\
        \midrule
        \textbf{GoogLeNet}         & \textbf{86.45} & CosineAnnealingWR \\
        \textbf{InceptionV3}       & \textbf{86.45} & ReduceLROnPlateau \\
        DenseNet          & 86.17 & OneCycleLR \\
        RegNet            & 85.96 & ExponentialLR \\
        FractalNet        & 85.55 & StepLR \\
        VisionTransformer & 85.55 & ReduceLROnPlateau \\
        SwinTransformer   & 85.46 & CosineAnnealingLR \\
        EfficientNet      & 85.12 & StepLR \\
        BagNet            & 84.69 & StepLR \\
        MobileNetV2       & 84.55 & CosineAnnealingLR \\
        ResNet            & 84.53 & LinearLR \\
        ICNet             & 84.49 & MultiStepLR \\
        AirNet            & 84.42 & MultiStepLR \\
        DPN               & 84.25 & OneCycleLR \\
        UNet2D            & 83.92 & CyclicLR \\
        ConvNeXt          & 83.73 & CosineAnnealingLR \\
        DarkNet           & 83.67 & ExponentialLR \\
        BayesianNet       & 83.60 & PolynomialLR \\
        SqueezeNet        & 81.55 & CyclicLR \\
        MobileNetV3       & 81.31 & CyclicLR \\
        \bottomrule
    \end{tabular}
\end{table}

\subsection{Architecture--Scheduler Interactions}

No single scheduler is best for all architectures
(Table~\ref{table:per_arch}). The two inception-based networks reach
the same peak (86.45\%) under different schedulers:
CosineAnnealingWarmRestarts for GoogLeNet and ReduceLROnPlateau for
InceptionV3. The best runs of the two transformers use
CosineAnnealingLR (SwinTransformer) and ReduceLROnPlateau
(VisionTransformer). UNet2D, SqueezeNet and MobileNetV3 all peak with
CyclicLR, whereas MobileNetV2, EfficientNet and ResNet peak with
CosineAnnealingLR, StepLR and LinearLR respectively. These groupings
rest on two or three architectures each and on single best runs, so
they are indicative only.

\subsection{Discussion}

For screening pipelines that evaluate many architectures with a
short, fixed budget, the results suggest two practical points. First,
the scheduler mostly affects how reliably a short run reaches good
accuracy rather than the best accuracy reachable, so a screen is
mainly useful for ruling out configurations that train unreliably.
Second, how a scheduler is stepped matters: ReduceLROnPlateau stepped
per batch on the training loss has the lowest mean but still produced
the best run for two architectures.

\section{Limitations}
\label{sec:limitations}

Several limitations remain. First, evaluation is limited to five
training epochs due to computational constraints, and rankings after
full-length training were not measured; they may differ, particularly
for schedulers with long warm-up phases such as OneCycleLR. Some
configurations are only partly expressed within five epochs: the
MultiStepLR milestones $\{5,10\}$ lie at or beyond the end of training,
and CosineAnnealingLR with $T_{\max} \in \{10, 20\}$ covers only the
first part of its cosine curve. Second, the scheduler catalogue does not
cover all strategies, such as learned schedules, warmup-cosine
combinations common in large-scale transformer training, or
schedule-free optimisation.
Third, all experiments use SGD with fixed $\text{lr} = 0.01$; interactions
with Adam or AdamW optimizers are not explored. Fourth, the evaluation
is restricted to CIFAR-10; generalization to ImageNet-scale datasets,
detection, segmentation, and regression tasks remains to be
investigated. Fifth, ReduceLROnPlateau is stepped per batch on the
training loss instead of per epoch on a validation metric, which
differs from its usual use and limits what its results say about the
scheduler itself. Sixth, per-architecture results are maxima over runs
and all results come from single runs without confidence intervals, so
differences of less than about one percentage point should not be
interpreted. Finally, ten of the 30 architectures could not be
attributed individually, and architectures designed for
non-classification tasks (UNet2D, Diffuser, ICNet) may be at a
disadvantage on CIFAR-10.

\section{Conclusion}
\label{sec:conclusion}

We presented a systematic large-scale evaluation of learning rate
scheduling strategies across 30 neural network architectures. By
automating scheduler injection via programmatic source-code modification, we
contributed and evaluated 3,938 model variants on CIFAR-10 within
the LEMUR nn-dataset, with the best configuration reaching 86.45\%
top-1 accuracy and 237 variants exceeding 80\%.

Our key findings are fourfold. First, no single scheduler is best
across all architectures: the best runs of the 20 attributed
architectures are spread over all ten scheduler families. Second,
scheduler family shifts mean five-epoch accuracy by about ten
percentage points (45.04--55.20\%) but the best result per family by
only about two (84.25--86.45\%). Third, ReduceLROnPlateau, stepped per
batch, has the lowest mean yet produced the best run for two
architectures, so the way a scheduler is stepped matters in
short-budget screening. Fourth, all non-zero weight-decay values have a
higher mean than $\lambda = 0$, an observation that remains to be
confirmed with balanced sample sizes.

Future work will validate these five-epoch findings against full-length
training and extend the framework
to multi-objective scheduling (accuracy vs.\ convergence speed),
incorporate warm-up combinations and adaptive optimizers such as
Adam and AdamW, and generalise to ImageNet-scale classification,
object detection, and segmentation.

{
    \small
    \bibliographystyle{ieeenat_fullname}
    \bibliography{bibmain}
}

\end{document}